\documentclass[conference]{IEEEtran}
\usepackage{cite}
\usepackage{amsmath,amssymb,amsfonts}
\usepackage{algorithmic}
\usepackage{algorithm2e}
\usepackage{algorithm}
\LinesNumbered 
\usepackage{graphicx}
\usepackage{textcomp}
\usepackage{xcolor}
\usepackage[numbers]{natbib}
\def\BibTeX{{\rm B\kern-.05em{\sc i\kern-.025em b}\kern-.08em
    T\kern-.1667em\lower.7ex\hbox{E}\kern-.125emX}}
\usepackage[breaklinks=true]{hyperref}
\usepackage{breakurl}

\begin{document}

\title{Causal-discovery-based root-cause analysis and its application in time-series prediction error diagnosis}


\author{\IEEEauthorblockN{1\textsuperscript{st} Hiroshi Yokoyama}
\IEEEauthorblockA{\textit{Graduate School of Interdisciplinary Science and Engineering in Health Systems} \\
\textit{Okayama University}\\
Okayama, Japan\\
h-yokoyama@okayama-u.ac.jp}
\and
\IEEEauthorblockN{2\textsuperscript{nd} Ryusei Shingaki}
\IEEEauthorblockA{\textit{System AI Laboratory} \\
\textit{Toshiba Corporation}\\
Kanagawa, Japan \\
ryusei1.shingaki@toshiba.co.jp}
\and
\IEEEauthorblockN{3\textsuperscript{rd} Kaneharu Nishino}
\IEEEauthorblockA{\textit{System AI Laboratory} \\
\textit{Toshiba Corporation}\\
Kanagawa, Japan \\
kaneharu1.nishino@toshiba.co.jp}
\and
\IEEEauthorblockN{4\textsuperscript{th} Shohei Shimizu}
\IEEEauthorblockA{\textit
{Faculty of Data Science} \\
\textit{Shiga University}\\
Shiga, Japan \\
shohei-shimizu@biwako.shiga-u.ac.jp}
\and
\IEEEauthorblockN{5\textsuperscript{th} Thong Pham}
\IEEEauthorblockA{\textit
{Faculty of Data Science} \\
\textit{Shiga University}\\
Shiga, Japan \\
thong-pham@biwako.shiga-u.ac.jp}
}


\maketitle

\begin{abstract}%
Recent rapid advancements of machine learning have greatly enhanced the accuracy of prediction models, but most models remain “black boxes”, making prediction error diagnosis challenging, especially with outliers. This lack of transparency hinders trust and reliability in industrial applications. Heuristic attribution methods, while helpful, often fail to capture true causal relationships, leading to inaccurate error attributions. Various root-cause analysis methods have been developed using Shapley values, yet they typically require predefined causal graphs, limiting their applicability for prediction errors in machine learning models. To address these limitations, we introduce the Causal-Discovery-based Root-Cause Analysis (CD-RCA) method that estimates causal relationships between the prediction error and the explanatory variables, without needing a pre-defined causal graph. By simulating synthetic error data, CD-RCA can identify variable contributions to outliers in prediction errors by Shapley values. Extensive experiments show CD-RCA outperforms current heuristic attribution methods.
\end{abstract}


\section{Introduction}
With the recent development of artificial intelligence (AI) technology based on various machine learning (ML) architectures, the predictive capability of ML methods has drastically improved. While applying ML-based prediction techniques in an industrial situation has attracted growing attention, the prediction mechanisms in most modern ML architectures (e.g., deep neural networks) are close to black box systems~\cite{10.1145/3236009,deep_learning_1}. When there is something unexpectedly happened that lead to an unusually bad performance of the ML system, i.e., an outlier in the prediction error, the diagnosis of the reason for this outlier, e.g., how much each explanatory variable contributed to the outlierness, is of paramount importance. It might not only lead to preventions of future catastrophic outliers, but also provide clues to improve the overall reliability, safeness, and trustworthiness of the ML system~\cite{trust_AI_survey_1,trust_AI_survey_3}. However, the black-box nature of most modern ML architectures makes the diagnosis not straightforward, and in some cases near impossible, to perform manually by a human. Therefore, establishing an eXplainable AI (XAI) method to automatically diagnose outliers in prediction errors would be crucial to improve the development of AI applications for practical and industrial situations~\cite{trust_AI_survey_2,Hassija2023InterpretingBM}. However, current methods for understanding the root cause of outliers in prediction errors is still lacking. 

In the context of XAI studies, various heuristic analysis methods for detecting the attributive variables in ML models for prediction error have been proposed~\cite{LiME_origin, ide_2021_lc, deng_aaai_2021,attribution_survey,ide2023,deng_tpami_2024}. However, since these methods do not consider the causal relationships between the prediction error, the target variable, and the explanatory variables in the ML method, existing heuristic attribution methods do not necessarily reflect the causality in the generating process of the prediction error \cite{Huang2024, Ma2020_PMLR}, and thus might lead to erroneous attributions, as will be shown in Fig.~\ref{fig:illustrative}.

Recently, root-cause analysis (RCA) that utilizes causality through various variants of the Shapley values (SV)~\cite{Shapley} have been extensively studied, with various method proposed~\cite{Heskes_neurips_2020,do_shapley_2022,pmlr-v162-budhathoki22a,janzing24a}.  However, these methods are focused on the RCA of abrupt changes in target observations, rather than that of the prediction error from some black-box prediction model. Furthermore, to directly apply these methods to the attribution problem of prediction errors, one needs a causal graph that represents the causal relationships between the explanatory variable, the target variable, and the prediction error. Since the estimation of such a causal graph has not been considered in those works, it remains unclear whether these methods can be applied in XAI.

To address the above issues, we propose Causal-Discovery-based Root-Cause Analysis (CD-RCA), a novel XAI method that directly considers the causal processes behind prediction errors in machine learning-based methods. The key idea in our approach is to approximate the unobtainable causal relationships between the explanatory variables, the target variable, and the prediction error of a black-box prediction model $f$ with a tractable causal structural model. This surrogate model allows us to generate synthetic samples of the prediction errors that approximately preserve the causality in the actual generating process of the observed prediction errors. This allows us to ask what would the prediction error look like if an explanatory variable $X_p$ had followed its normal causal mechanism? Roughly speaking, such a counterfactual prediction error can be approximately generated from the surrogate causal model by randomizing the exogenous noise at $X_p$. Comparing this counterfactual prediction error with the actual outlier prediction error would give us an idea about the contribution of $X_p$. Specifically, the attribution of each variable to an outlier of the prediction error can be quantified by employing a version of the Shapley values, proposed by~\cite{pmlr-v162-budhathoki22a}. At a high level, our main contributions are:
\begin{itemize}
\item We propose the CD-RCA method for model-agnostic prediction error attribution, and provide an adaption of the method to time-series data. Our method is an extension of~\cite{pmlr-v162-budhathoki22a}, in the sense that, unlike their method that require a known causal graph, our method does not assume any causal graph is known, and proceeds to estimate the causal graph between the explanatory variables, the target variable, and the prediction error. We show through extensive simulations that CD-RCA outperforms existing methods in identifying the root cause of outliers in prediction errors.
\item Through sensitivity analysis, we identified various previously-unknown patterns in which the Shapley values might fail to capture the correct attributions. Namely, CD-RCA  might encounter difficulties in detecting the root cause variable when: (1) the amplitude of the anomalous noise is less than the mean of the exogenous noise, or (2) there is no causal effect of the root cause variable on the target variable. These results might have independent interests, as they provide clues for investigating other attribution methods that incorporate variants of the Shapley values.
\end{itemize}

The organization of the paper is as follows. We introduce our proposed method in Section~\ref{sec_proposed_method}. We provide an illustrative example in Section~\ref{sec:sub_illustrative} and show the sensitivity analysis results in Sections~\ref{section_ex_sim1} and ~\ref{section_ex_sim2}. A case study using real-world data is given in Section~\ref{sec:real_world}. Concluding remarks are given in Section~\ref{sec:conclusion}. 
The code used to reconstruct all the results in this paper was implemented
using Python and will be made publicly available.

\section{Root-cause analysis of prediction error outliers}
\label{sec_proposed_method}
We explain the basic setup for non-time-series data in Section~\ref{sub_sec:setup}. We discuss how to measure the outlier score of prediction errors in Section~\ref{sub_sec:outlier_score}, and a general framework to calculate the attribution of each covariate to the outlier score in Section~\ref{sub_sec:general_framework}. In Section~\ref{sub_sec:timeseries_case}, we extend the framework to time-series data. In Section~\ref{sub_sec:framework_discuss}, we discuss some modelling and computational aspects of our framework.
\subsection{Setup}\label{sub_sec:setup}
Suppose that we observe $n$ samples from the joint distribution of the target variable $Y$, the covariates $X_{1}$, $\ldots$, $X_{d}$, and the prediction error $r = Y - \hat{Y}$, where $\hat{Y}$ is obtained from  from some black-box prediction model. Denote these samples as $D_{\text{train}} = (X_{i,1},\ldots,X_{i,d},Y_i, r_i)_{i=1}^{n}$. Assume that we want to use $D_{\text{train}}$ to learn a model to measure the outlierness of $r^{*}$ in some target sample $((x_{p}^*)_{p=1}^{d},y^{*},r^{*})$ from the joint distribution of $(X_{1}$, $\ldots$, $X_{d},Y,r)$. Note that this sample is not necessarily included in $D_{\text{train}}$. We also want to quantify the contribution of each $X_p$ ($p = 1,\ldots, d$) and $Y$ to the outlierness of $r^{*}$. To streamline the notation, we will sometimes use $X_{d+1}$, $X_{i,d+1}$, and $x_{d+1}^{*}$ in place of $Y$, $Y_i$, and $y^{*}$, respectively, and $X_{d+2}$, $X_{i,d+2}$, and $x_{d+2}^{*}$ in place of $r$, $r_i$, and $r^{*}$, respectively. 
\subsection{Outlier scores}\label{sub_sec:outlier_score}
For measuring the outlierness of $r^{*}$, we use the following information-theoretic score $S(r^{*})$~\cite{pmlr-v162-budhathoki22a}. 
\begin{equation}
S(r^{*}) =  -\log \mathbb{P}\{\tau(r) \ge \tau(r^{*})\},\label{eq:score_def}
\end{equation} 
where $\tau:\mathbb{R}\longrightarrow\mathbb{R}$ is some transformation.
The outlier score $S(r^{*})$ can be calculated, for example, by using the empirical distribution $(r_i)_{i=1}^{n}$ of $r$. This outlier score includes the familiar $z$-score when $\tau(r) = |r - \mu_{r}|/\sigma_{r}$ with $\mu_{r}$ and $\sigma_{r}$ being the mean and standard deviation of the distribution of $r$, respectively. 

This outlier score measures the outlierness of the value $r^{*}$ by the probability of the event $\tau(r) \ge \tau(r^{*})$. When this probability is small, which means that the event is rare, the outlier score will be large. For example, considering the transformation $\tau(r) = |r - \mu_{r}|/\sigma_{r}$, a $r^{*}$ that is unusually large when taken into accounts the mean and the variance of the distribution will give a large value of $\tau(r^{*})$, thus leads to a large $S(r^{*})$. Such an outlier $r^{*}$ can be caused by either outlier values in one or multiple covariates in $X_1$, $\ldots$, $X_d$, and/or by an outlier value of the target variable $Y$. In the next section, we aim to \emph{quantify the contribution $\phi(p)$ of $X_p$ to the outlier score $S(r^*)$}, in such a way that reflects \emph{the degree of causation}~\cite{Halpern2015-HALGCA}: the more $X_p$ causes the outlier $r^{*}$, the larger the value of $\phi(p)$ should be.

\subsection{The general framework}\label{sub_sec:general_framework}
At the core of our framework is the approximation of the unattainable causal relationships between the variables $X_1,\ldots, X_{d}$, $Y$ and $r$ by the following tractable surrogate causal model, which is known as additive noise models in the causality literature~\cite{hoyer_2009,judea_2009_book}:
\begin{equation}
X_p = g_{p}((X_{q})_{q \in pa(p)}) + N_{p},\ p =1,\ldots,d+2,\label{eq:functional_causal_model}
\end{equation}
where $pa(p)$ is the index set of the parents of $X_p$, and $N_p$ is the exogenous noise random variable at $X_p$.

We build this surrogate causal model in three sub-steps. Firstly, we learn a causal graph $G$ between $(X_1,\ldots, X_d,Y,r)$ by some causal discovery algorithm from the observational data. Secondly, we learn a parametric model $g_p$ for each variable $X_p$, conditioning on its parents $pa(p)$. Based on the fitted values of the model, we then calculate the observed noise value of each variable for each sample: $n_{i,p} = X_{i,p} - \widehat{X_{i,p}}$ for $p = 1$, $\ldots$, $d+2$ and $i = 1$, $\ldots$, $n$. Finally, we fit a parametric distribution for $N_p$ based on the samples $n_{i,p}$. For an exogenous variable $X_p$ whose parent set $pa(p)$ is empty, we fit a parametric distribution directly on $X_{i,p}$. By simulating a noise vector $(N_p)_{p=1}^{d+2}$ from the learned noise distributions, one can generate a sample $(X_1,\ldots, X_d,Y,r)$ using~(\ref{eq:functional_causal_model}). The noises in the target sample $(y^{*},(x_{p}^*)_{p=1}^{d},r^{*})$ can also be calculated from the causal graph and the learned model. In particular, the noise at $X_p$ in the target sample is
\begin{equation}
n_{p}^{*}= x_{p}^{*} - g_{p}((x_{q}^{*})_{q \in pa(p)}),\ p=1,\ldots,d+2,\label{eq:noise_calculation}
\end{equation}
where $g_{p}((x_{q}^{*})_{q \in pa(p)}) = 0$ if $pa(p) = \emptyset$.

To calculate the attributions of each variable to $S(r^{*})$, we employ the Shapley values proposed by~\cite{pmlr-v162-budhathoki22a}. Denote $S(r^{*} \mid U)$ the counterfactual outlier score of $r^{*}$ calculated from simulated data in which the noise value of a variable $X_{q}$ with $q\notin U$ is fixed at its observed value $n_{q}^{*}$ in the target sample, while we randomize the noise $N_{k}$ of $X_{k}$ for $k \in U$. This randomization can be done by sampling from the joint empirical distribution or the learned joint parametric noise distribution of $(N_k)_{k\in U}$. The value of $S(r^{*} \mid U)$ is counterfactual in the sense that it is the hypothetical value of $S(r^{*})$ had the noises of the variables in $U$ been randomized, i.e., not at their observed values in the target sample. Thus, the randomization has a whitening effect: the contributions of the variables $(X_{k})_{k\in U}$ to the outlierness of $r^{*}$ has been whitened out in $S(r^{*} | U)$. 

Suppose that we have already randomized the noise values of variables in some conditioning set $I \subseteq \{1,\ldots, d+2\}\setminus \{p\}$. This means that the contributions of the variables $(X_k)_{k\in I}$ to $S(r^{*})$ have been whitened out. The change in the outlierness of $r^{*}$ when we further randomize $N_{p}$ comparing with the current baseline outlierness of $r_i$, i.e., the difference 
\begin{equation}
\phi(p|I) = S(r^{*} \mid I\cup \{p\}) - S(r^{*} \mid I),\label{eq:conditioned_contribution}
\end{equation}
expresses the contribution of $X_p$ to $S(r^{*})$ when the contributions of $(X_k)_{k\in I}$ have been whitened out. Thus, $\phi(p | I)$ untangles the contribution of $X_p$ from the contributions of $(X_k)_{k\in I}$. 

Since this contribution depends on the conditioning set $I$, one can remove this dependency by defining the final, unconditional attribution of $X_p$ as a weighted average of all $\phi(p | I)$:
\begin{equation}
\phi(p) = \sum_{I \subseteq \{1,\ldots,d+2\}\setminus \{p\}} w_{I}\phi(p | I),
\end{equation}
where $w_{I} = (d+2)^{-1}{\binom{d+1}{|I|}}^{-1}$ and $\sum_{I \subseteq \{1,\ldots,d+2\}\setminus \{p\}}w_{I} = 1$.

This particular choice of the weights makes $\phi(p)$ precisely a Shapley value~\cite{Shapley}, and thus the attribution $\phi(p)$ possesses some desirable properties from a game-theoretic perspective, including the following \emph{efficiency property}:
\begin{equation} 
S(r_i) = \sum_{p=1}^{d+2}\phi(p).
\end{equation}
Note that this set of weights is the \emph{unique} set that can achieve efficiency (under the presence of some other desirable properties)~\cite{Grabisch}. We summarize our framework in Algorithm~\ref{algo:general_framework}. See Appendix~\ref{sec:related_works} for a discussion on related works. 


\begin{algorithm}[hbt!]
\caption{Causal-Discovery-based Root-Cause Analysis (CD-RCA)}\label{algo:general_framework}
 \KwIn{Training data $D_{\text{train}} = (X_{i,1},\ldots,X_{i,d},Y_i, r_i)_{i=1}^{n}$, target outlier sample $((x_{p}^*)_{p=1}^{d},y^{*},r^{*})$, a transformation $\tau$ }
 \KwOut{Outlier score $S(r^{*})$, attributions $(\phi(p))_{p=1}^{d+2}$}
 
  Estimate a causal graph $G$ from $D_{\text{train}}$\; 
  
 Learn parametric models for $g_p$ and the distributions of $N_p$ in~(\ref{eq:functional_causal_model})\;

 Calculate $S(r^{*})$ by~(\ref{eq:score_def}) using either observed data $(r_i)_{i=1}^{n}$ or synthetic data from the learned generative model\;
  
 Calculate noises in the target sample by~(\ref{eq:noise_calculation})
 
 \For{each $p$ in $1,\ldots, d+2$}{
      Initialize $\phi(p) \gets 0$\;
      
     \For{each set $I \in 2^{\{1,\ldots,\ d+2\}\setminus\{p\}}$}{
         calculate $\phi(p|I)$ using~(\ref{eq:conditioned_contribution}) by generating synthetic data \;
         
     $\phi(p) \gets \phi(p) +  (d+2)^{-1}\binom{d+1}{|I|}^{-1}\phi(p|I)$
     }
}
\end{algorithm}

\subsection{The case of time-series data}\label{sub_sec:timeseries_case}
Suppose that, instead of $n$ i.i.d. samples, we are given time-series data $D_{\text{train}} = (X_{t,1},\ldots,X_{t,d},Y_t, r_t)_{t=1}^{T}$ and a target outlier sample $((x_{t^{*},p})_{p=1}^{d},y_{t^*},r_{t^*})$ at some time-step $t^{*}$. We want to calculate the outlier score $S(r_{t^*})$ and the attributions $\phi(p)$ to this score. The general framework in Section~\ref{sub_sec:general_framework} can be adapted to this case.

For time-series data, we assume the following stationary
discrete-time autoregressive process with additive noises:
\begin{equation}
X_{t,p} = g_{p}((X_{t,q})_{q \in pa(t,p)}) + N_{t,p},\ p =1,\ldots,d+2,\label{eq:time_series_model}
\end{equation}
where $pa(t,p) \subseteq \{X_{t',q} \mid t' \le t\}$ is the set of parents of $X_{t,p}$ and $N_{t,p}$ is the noise at the variable $X_{t,p}$. Comparing with the general time-series functional causal model in~\cite{pcmci}, this model is a special case where the noise at each variable is additive.

The time-series data from this model can be approximately converted to a non-time series data as follows. Define $\tau_{max}$ as the order of the time-series. This means that, a variable $X_{t,p}$ is not a parent of any variable $X_{t + \tau,q}$ with $\tau > \tau_{max}$. This implies that $X_{t,p}$ and $X_{t+\tau_{max} + 1,p}$ can be approximately treated as independent samples from a same variable $X_{p}$. Therefore, we can create a new non-time-series dataset of $(d+2)\times (\tau_{max} + 1)$ variables: $X_{1}$ at lag $0$, $\ldots$, $X_{d+2}$ at lag $0$, $X_{1}$ at lag $1$, $\ldots$, $X_{d+2}$ at lag $1$, $\ldots$, $X_{1}$ at lag $\tau_{max}$, $\ldots$, $X_{d+2}$ at lag $\tau_{max}$ to approximate the original time-series. The framework in Algorithm~\ref{algo:general_framework} can then be applied to this dataset.

With this converted time-lagged dataset, Algorithm~\ref{algo:general_framework} will output the attribution of a variable $X_p$ at multiple time lags: $\phi(X_{p} \text{ at lag $0$})$, $\phi(X_{p} \text{ at lag $1$})$, $\ldots$, $\phi(X_{p} \text{ at lag $\tau_{max}$})$. We will use their sum as the final attribution of $X_p$: 
\begin{equation}
\phi(p) = \sum_{\tau = 0}^{\tau_{max}} \phi(X_{p} \text{ at lag $\tau$}).
\end{equation}
\subsection{Discussion}\label{sub_sec:framework_discuss}

\subsubsection{The choice of the causal discovery algorithm}
For non-time-series data, one can use constraint-based algorithms such as the Peter-Clark (PC) algorithm~\cite{peter_spirtes_book}, or methods based on functional causal models such as the direct method for learning a linear non-Gaussian acyclic model (DirectLiNGAM)  ~\cite{Shimizu2011_DirectLiNGAM} to estimate the causal graph. 

For time-series data, there are two ways to estimate the causal graph. The first way is to use time-series specific algorithms on the original time-series data. Some examples are constraint-based time-series algorithm PC algorithm with momentary conditional independence (PCMCI)~\cite{pcmci} and its bootstrap aggregation extension~\cite{PCMCIboot2024}, and the time-series algorithm vector-autoregressive LiNGAM (VARLiNGAM) that is based on functional causal models~\cite{var_lingam}. Alternatively, one can apply the PC or DirectLiNGAM algorithm on the converted lagged data.

\subsubsection{The choice of regression models and parametric noise distributions}
We use the \texttt{dowhy-gcm} package~\cite{dowhy_gcm} to learn the regression model for $g_p$ and parametric distribution for the noise $N_p$. This implementation includes a wide range of regression methods as well as rich families of parametric distributions, and cross-validation methods for selecting the optimal ones. We use the default options of the package in our experiments.



\section{Simulation study}
We will compare our method with existing approaches in detecting root causes for prediction errors. We also investigate the limitations and sensitivity of our proposed approach.

For baselines of anomaly detection methods used in comparisons, we applied the following methods: Local Interpretable Model-agnostic Explanations (LIME)~\cite{LiME_origin}, Expected Integrated Gradien (EIG)~\cite{deng_aaai_2021, deng_tpami_2024},  Likelihood Compensation (LC)~\cite{ide_2021_lc}, Generative Perturbation Analysis (GPA)~\cite{ide2023}, and z-score. The implementations of the baselines are adapted from Python scripts of prior work~\cite{ide2023}. See~\cite{LiME_origin, ide_2021_lc, ide2023} for more details of each baseline method. These baselines are widely used for RCA in machine learning studies. These methods do not consider the causal relationship between prediction error and covariates (i.e., these methods defined the root cause based on the ``correlation'' between prediction error and covariates).

In Section~\ref{sec:sub_illustrative}, we provide an example to showcase the working of our method CD-RCA. We then offer various sensitivity analyses of CD-RCA in Sections~\ref{section_ex_sim1} and ~\ref{section_ex_sim2}. 

\subsection{An illustrative example}
\label{sec:sub_illustrative}

\begin{figure}
\centering\includegraphics[width=.85\linewidth]{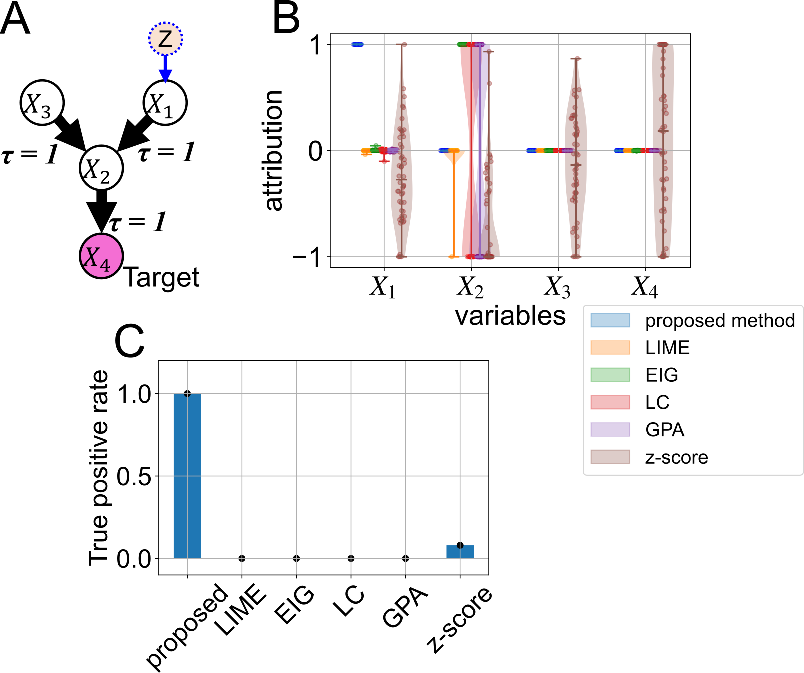}
    \vspace{-5pt}
    \caption{\textbf{Root cause detection for prediction error outlier in time-series data.} (A) The true causal graph. The prediction target variable is $Y = X_4$. The root-cause for the outlier sample is $X_1$. (B) Normalized attribution of each variable provided by each method. The probability densities in the violin plots were calculated using a kernel density method based on attribution values from $50$ trials. The error bar indicates the maximum and minimum values of the attributions. (C) True positive rate in identifying $X_1$ as the root cause of each method.}
    \label{fig:illustrative}
    \vspace{-10pt}
\end{figure}

We consider synthetic time-series data generated from the model in~(\ref{eq:time_series_model}). In this example, there are three covariates $X_1$, $X_2$, $X_3$, and a target variable $Y = X_4$. The causal relationships between them are described in Fig.~\ref{fig:illustrative}A. Twenty thousand samples of $((X_p)_{p=1}^3,Y)$ were generated from the causal model described in~(\ref{eq:illustrative}).

Using the first $10000$ sample, we trained a time-series prediction model proposed in the prior work\cite{Runge2015}. This trained model would be subsequently used as the black-box prediction model $f$. We then augmented each sample of $((X_p)_{p=1}^3,Y)$ in the last $10000$ samples with the corresponding prediction error $r$ from the prediction model. The final samples are of the form $((X_p)_{p=1}^3,Y,r)$, and constitute our training data $D_{train}$ for Algorithm~\ref{algo:general_framework}.    

To generate a target outlier sample, we first generated a new time-series of $((X_p)_{p=1}^3,Y)$ from~(\ref{eq:illustrative}). When generating the sample at the target sample $T$, we first generated a clean $X_1$ based on~(\ref{eq:illustrative}), then added an outlier noise $Z = 20$ to the result. We then used the contaminated value of $X_1$, together with $X_3$, to generate $X_2$, and then $Y$, based on~(\ref{eq:illustrative}). This contaminated sample is then put into the prediction model $f$ to obtain the prediction error $r$. Since the value of $X_1$ was unusually large, the value of $r$ was unusually large, too, i.e., $X_1$ is the root cause for the outlier value of $r$. The augmented sample of $((X_p)_{p=1}^3,Y,r)$ is then chosen as the target outlier sample. 

For estimating the causal graph from $D_{train}$ in CD-RCA, we use the PCMCI algorithm with bootstrap aggregation~\cite{PCMCIboot2024} with number of bootstrap samples set at $500$ as the causal discovery algorithm in Algorithm~\ref{algo:general_framework}. For each run of Algorithm~\ref{algo:general_framework}, we obtained one set of attribution values. When calculating the anomaly score and attribution values, we set the sampling number as $N=10,000$ to obtain joint empirical distributions.
Due to randomness in line 8 of the algorithm, we repeated Algorithm~\ref{algo:general_framework} $50$ times with different 50 target anomaly samples. As a result, we can obtain $50$ sets of attribution values.

For RCA with baseline methods, we applied five baselines: LIME, EIG, LC, GPA and z-score to these $50$ target anomaly samples. 

Fig.~\ref{fig:illustrative}B shows the normalized attribution of each variable over $50$ sets of attribution values. All baselines wrongly gave the largest attribution to any other variables (especially, $X_2$) than exact root cause $X_1$, while our CD-RCA successfully gave $X_1$ the highest attribution. Fig.~\ref{fig:illustrative}C shows the true positive rate in successfully identifying $X_1$ as the root cause. Moreover, the baselines could not identify $X_1$ as their true positive rates are almost $0$, while CD-RCA successfully identified $X_1$ as the root-cause of the outlier. 

To \emph{predict} the value of $Y$, the value of $X_2$ alone is enough, while in the absence of $X_2$ both value of $X_1$ and $X_3$ are needed. In a sense, this implies that $X_2$ might be the most important variable in predicting $Y$. This is the reason most baselines picked out $X_2$ as the root cause, and this example shows that focusing only on powers in predicting $X_4$ is not enough. Since CD-RCA considers the causal relationships between all variables, it could successfully detect the root cause as $X_1$.

\subsection{Effects of outlier magnitude and average total effect on CD-RCA}
\label{section_ex_sim1}
Since the attribution $\phi(p)$ in CD-RCA is based on the Shapley value proposed in prior study~\citep{pmlr-v162-budhathoki22a}, the reliability of CD-RCA would be directly affected by the RCA performance of this particular version of the Shapley value, which surprisingly has not been investigated in previous works.
Therefore, before evaluating our proposed CD-RCA method, the reliability of Shapley values should be confirmed since our method is directly affected by the performance of the Shapley value based RCA algorithm~\citep{pmlr-v162-budhathoki22a}. 

In this experiment, we evaluated the sensitivity of the RCA part to changes in amplitudes of anomalous noises and changes in causal mechanisms. 
To isolate the RCA performance from other parts of CD-RCA, although we still execute line 2 of Algorithm~\ref{algo:general_framework} to learn the generative model from the observed data, we do not estimate the causal graph, i.e., line 1 in Algorithm~\ref{algo:general_framework}, and use the true causal graph. Furthermore, we do not include prediction errors for directly assessing the RCA-performance of Shapley value based method ~\citep{pmlr-v162-budhathoki22a}. However, we expect that the sensitivity result of this experiment would also apply to the full version of CD-RCA, a point that we confirmed in Section~\ref{section_ex_sim2}. Instead of prediction errors, another variable was chosen to pick out an outlier sample and measure its outlier score.


The causal mechanisms for training data were based on the two models $F_1$ and $F_2$ in Eqs.~(\ref{eq_sim1_F1}) and~(\ref{eq_sim1_F2}), respectively. These models constructed with three covariates $X_p$ ($p=1,\ldots,3$) and a target variable $X_4$, where we will measure its outlier score and the attributions of the remaining variables to the score (see Appendix \ref{append1_sim1_equations}). 

For the target outlier sample, we will use different causal models compared to the training data. Specifically, the outlier samples corresponding to training data generated from $F_1$ and $F_2$ were generated from three models in Eqs. (\ref{eq_sim1_F1_testA})-- (\ref{eq_sim1_F1_testC}) and three models in Eqs.~(\ref{eq_sim1_F2_testA})--(\ref{eq_sim1_F2_testC}), respectively. These models for outlier samples have the same causal graphs as in $F_1$ and $F_2$, but with tunable coefficients $\beta$ of some edge and different locations of where we will add the anomalous noise $Z$. The amplitude of $Z$ can also be controlled. The root-cause $Z$ was added to one of the observational variables $X$ only at the $500$-th sample, and this sample is chosen to be the target outlier sample. See Appendix \ref{append1_sim1_equations} for more details.

By using the generated target outlier samples with the above procedures, 
the attribution $\phi$ in the target sample were evaluated by using CD-RCA with known causal graphs $F_1$ and $F_2$. Moreover, the sampling number is set to $N=50,000$ to obtain joint empirical distributions for calculating the attribution $\phi$.
For quantifying the root cause detection accuracy, the values of $\phi$ were repetitively estimated 50 times for all parameter conditions of $\beta$ and $Z$ for each model. After these estimations were finished, the RCA accuracy for each model was calculated for all parameter conditions. A successful detection was defined as when the variable with the maximum value of the attribution $\phi(p)$ is the same as where we added $Z$ in the outlier sample. 

\begin{figure}[h]
    \centering
    \includegraphics[width=.7\linewidth]{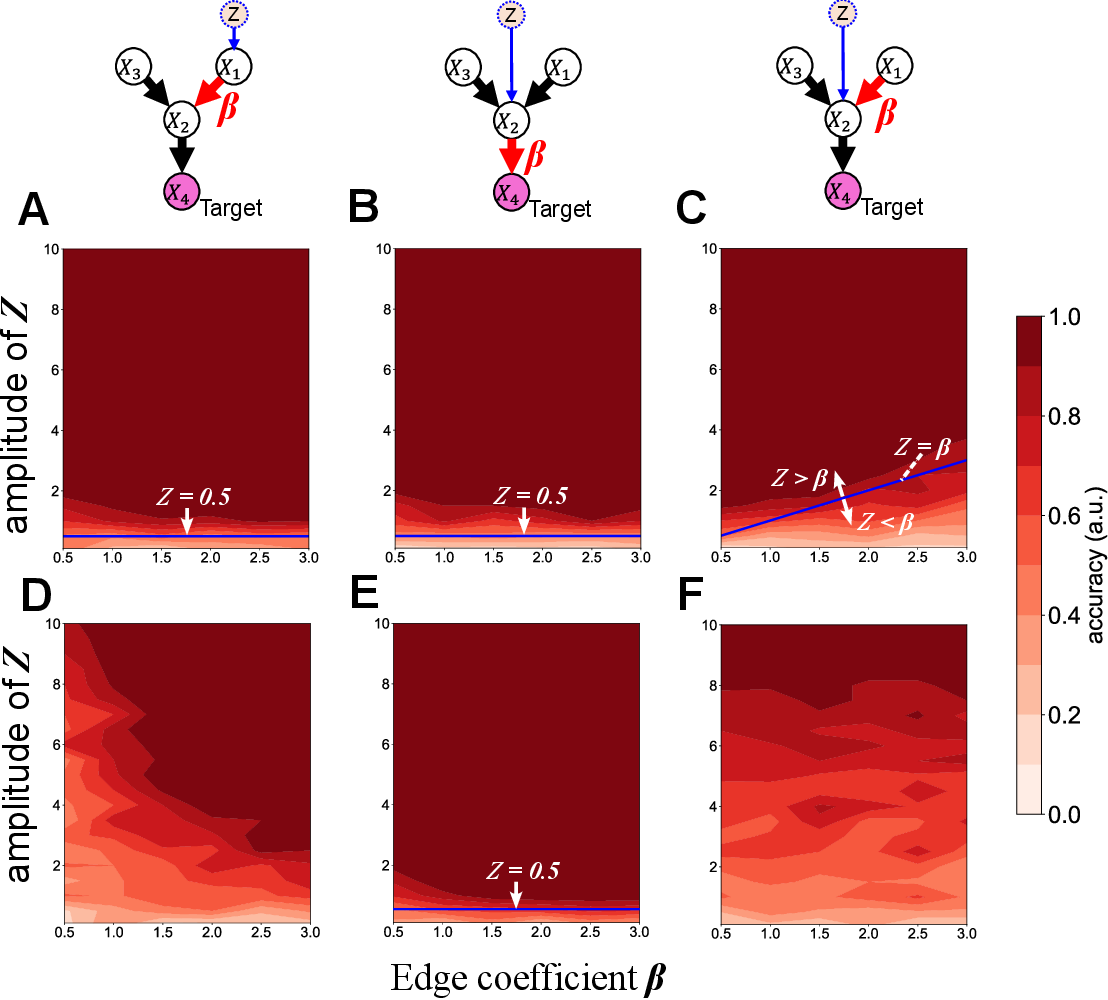}
    \caption{\textbf{Effects of outlier magnitude and average total effect to target variable on the performance of CD-RCA.} (A--C) The results of the root cause detection accuracy, relative to the changes in amplitude of $Z$ and graph edge weight $\beta$ in the model $F_1$. The causal graph diagram of each panel indicates the location of the exact root-cause ($Z$) and edge changes ($\beta$) for each simulation settings. (D--F) The results of the model $F_2$ in the same manner of (A--C). }
    \label{fig_results_noise_effect_time_series}
\end{figure}

The results with model $F_1$ are shown in Fig.~\ref{fig_results_noise_effect_time_series}A--C. Fig.~\ref{fig_results_noise_effect_time_series}A, B indicated that the RCA accuracy tends to be high regardless of the value of edge coefficient $\beta$ under the condition $Z \gg 0.5$. Since the exogenous noise $N_{p}$ at each $X_p$ follows the uniform distribution $\mathcal{U}(0, 1)$ with mean 0.5, these results suggested that the detection is successful when $Z \ge \mathbb{E}[N_{p}]$. However, the result in Fig.~\ref{fig_results_noise_effect_time_series}C showed a different tendency: the accuracy tends to be lower when $Z \le \beta$. One explanation is that the effect of $Z$ on $X_2$ becomes relatively lower than that of $X_1$ on $X_2$ when $Z\le\beta$, and thus $X_1$ tends to be wrongly identified as the root-cause in Fig.~\ref{fig_results_noise_effect_time_series}C.

\begin{figure}[h]
    \centering
    \includegraphics[width=.7\linewidth]{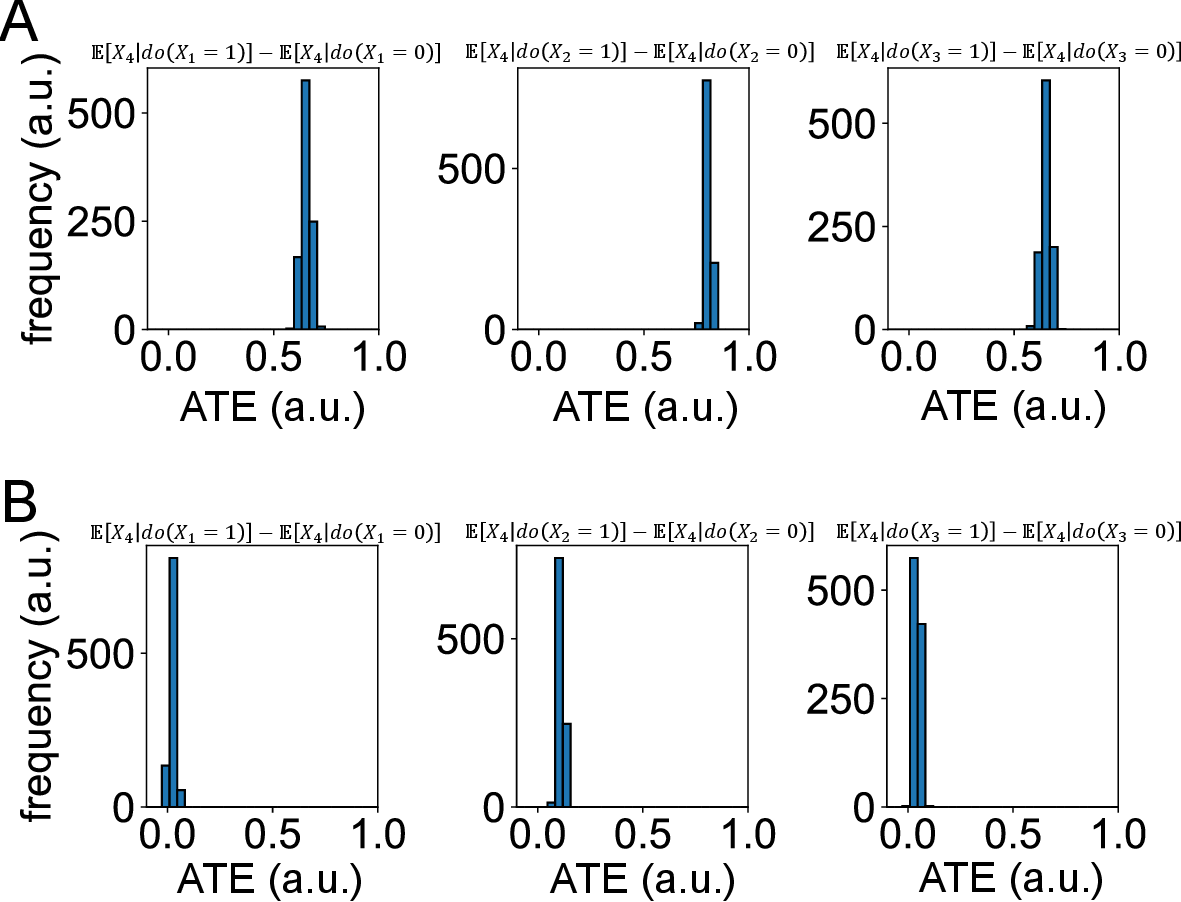}
    \caption{\textbf{Total effect to target variable $X_4$ in the models for training data.} A) Average total effect: ATE in the model $F_1$, from $X_1$ to $X_4$, from $X_2$ to $X_4$, and from $X_3$ to $X_4$, respectively. B) ATE results of $F_2$ in the same manner in $F_1$. These results were obtained from causal models $F_1$ and $F_2$ by the \texttt{dowhy-gcm} package.}
    \label{fig_sim1_total_effect}
\end{figure}

In contrast to the results with $F_1$ in Fig.~\ref{fig_results_noise_effect_time_series}A--C, the results with $F_2$ in Fig.~\ref{fig_results_noise_effect_time_series}D--F showed a different tendency. Even though the causal graph in $F_2$ is the same as in $F_1$, some edge coefficients in the causal graph were set to a smaller value relative to those in $F_1$ (see Eqs.~(\ref{eq_sim1_F1}) and~(\ref{eq_sim1_F2})). These differences were reflected in the differences in the average total effects (ATE) to the target variable $X_4$. Since the $ATE_{X_1 \rightarrow X_4}$ in $F_2$ , as can be seen from Fig.~\ref{fig_sim1_total_effect}B, were almost zero, i.e., $X_1$ has no causal effect on $X_4$, the high accuracy of RCA are not found unless the amplitude of $Z$ is large enough in Fig.~\ref{fig_results_noise_effect_time_series}D. The same phenomenon happened in Fig.~\ref{fig_results_noise_effect_time_series}F. Since the $ATE_{X_2 \rightarrow X_4}$ in $F_2$, as can be seen in  Fig.~\ref{fig_sim1_total_effect}B, were relatively higher than zero, the RCA performance in Fig.~\ref{fig_results_noise_effect_time_series}E showed similar tendency as that of Fig.~\ref{fig_results_noise_effect_time_series}B.

Considering the above results, the CD-RCA method might encounter difficulties in detecting the root cause variable when: (1) the amplitude of $Z$ is smaller than the observational noise $N$, or (2) the root cause variable has no causal effect on the target variables.

Since previous works have not examined the RCA performance of the Shapley value proposed in~\cite{pmlr-v162-budhathoki22a}, the findings in this section may be of independent interest.

\subsection{Effects of failed causal graph estimation on CD-RCA}
\label{section_ex_sim2}
In this simulation, we aimed to assess how errors in causal graph estimation affect RCA by the following simulation.

We use a stationary causal graph (see Appendix \ref{append_sim2}) that has a similar causal structure with the graph in Section~\ref{sec:sub_illustrative} to generate training data. The generated data consisted of three covariates $X_p$ ($p=1,\ldots,3$) and a target variable $Y=X_4$. When generating an outlier sample for RCA, an unobservable root cause $Z$ is added at $X_1$. The amplitude $Z$ is selected from $[0.5, 0.7, 0.9]$. The target variable of RCA is set as $X_4$. 

To control the degree of misspecification, we do not estimate the causal graph from training data. Instead, we deliberately modify the true causal graph and create an intervened graph. We then perform RCA with this intervened graph. Specifically, we alter the true causal graph in two ways: 
\begin{enumerate}
    \item Set the edge number $k$ ($k = [1,2,3]$) of the intervention
    \item While acyclicity is satisfied, add  $k$ edges to the true causal graph
\end{enumerate}
With an intervened causal graph, we perform RCA as follows.

\begin{enumerate}
    \item[a)] Generate 10 target outlier samples
    \item[b)] Generated an intervened causal graph
    \item[c)] Conduct RCA of each target outlier sample with the intervened causal graph
    \item[d)] Calculate the root cause detection accuracy based on RCA results obtained from 10 target samples 
    \item [e)] Repeat steps a) to d) 100 times. 
\end{enumerate}
When performing RCA in step c), the sampling number is set to $N=10,000$ to obtain joint empirical distributions to calculate the attribution value $\phi$.

The results of this experiment are shown in Table~\ref{tab_sensitivity}. 

\begin{table}[h]
    \centering
    \caption{RCA performance under graph intervention.}
    \begin{tabular}{c|c|c|c}
    $Z_{scale}$ & $k=1$ & $k=2$ & $k=3$ \\
    \hline\hline
    0.50 & 0.66 $\pm$ 0.07 & 0.66 $\pm$ 0.06 & 0.65 $\pm$ 0.07 \\
    \hline
    0.70 & 0.85 $\pm$ 0.05 & 0.85 $\pm$ 0.05 & 0.85 $\pm$ 0.05 \\
    \hline
    0.90 & 0.99 $\pm$ 0.01 & 0.99 $\pm$ 0.01 & 0.99 $\pm$ 0.01 \\
    \end{tabular}
    \label{tab_sensitivity}
\end{table}
Note that the value of each cell indicates the mean and standard deviation of RCA accuracy over 100 repetitions. $k$ stands for the number of intervention edges.   

As can be seen in the table, we found that root cause detection performance was more accurate when  $Z \gg \mathbb{E}[N_{back}] = 0.5$, where $\mathbb{E}[N_{back}]$ indicates the expectation of the observational noise. This is consistent with the findings in Section~\ref{section_ex_sim1}. Moreover, such tendencies are independent of $k$. Therefore, these results demonstrated the possibility that our proposed method would detect exact root causes even under the condition of graph misspecifications. However, future work should reveal the theoretical reasons for the results through further numerical and mathematical considerations.

\section{Case study: Root causes of outliers in predicting river inflow}\label{sec:real_world}
We apply our proposed CD-RCA method to a real-world dataset consists of river inflow volumes to the Taisetsu Dam, located around the Ishikari river in Hokkaido, Japan~\cite{dam_dataset1}. The objective of this experiment is to compare our proposed CD-RCA with existing baseline methods in identifying the root cause of prediction errors in river inflow forecasting using machine learning models. 

\begin{figure}[h]
    \centering
    \includegraphics[width=0.75\linewidth]{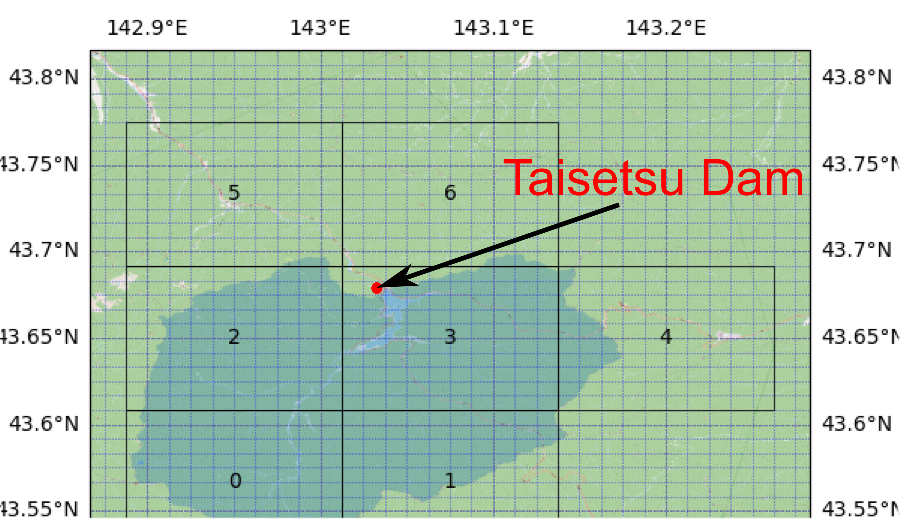}
    \caption{Map of the surrounding area of the Taisetsu dam. The numbering from $0$ to $6$ corresponds to the observation point $m$ of the covariates $q_{n}^{(m)}$ produced by the Sugawara tank model. The geographical information of these observation points is as follows. Upper stream of the Ishikari river: 0 and 2. Lower stream of the Ishikari river: 3. Other: 1,4,5 and 6.}
    \label{fig_dam_location}
\end{figure}

The dataset contains the volume of river inflow into the Taisetsu Dam obtained from January 1st, 2019, to December 31st, 2022, with 1-hour sampling intervals. Furthermore, we augmented the data with rainfall data, indexed from $m = 0$ to $m = 6$, within the catchment area of the Taisetsu Dam~\cite{dam_dataset2}. See Fig.~\ref{fig_dam_location} for the location of each rainfall observation point.

The entire system is highly non-linear due to various factors, including complex geography, varying precipitation/evaporation rates, sophisticated water infiltration into different types of land surfaces, and hard-to-measure groundwater systems.

The Sugawara tank model~\cite{Sugawara1961OnTA, Sato2023_TANK} was applied to pre-process the data. This is a hidden variable model that attempts to model the physics of water infiltrations and water retention. The Sugawara tank model generates three variables, indexed by $n = 0$ to $n = 2$, for each rainfall observation point. Given that we use rainfall data at seven observation points, the Sugawara tank model produces a total of 21 variables, denoted as $q_{n}^{(m)}$ ($m = 0,\ldots,6$ and $n=0,\ldots,2$). These variables serve as covariates for forecasting river inflow at the Taisetsu Dam.

We conduct the RCA experiments based on the following three steps. In the first steps, we first construct the time-series forecasting model for river inflow in the dam based on the Light-GBM algorithm. This model forecasts the river inflow one hour later in this dam, based on current and previous observations (including the river inflow and soil water accumulations,i.e., variables produced by the 3-TANK model, in seven observed points) from five-hour intervals, with one-hour sampling intervals. To learn and evaluate the prediction error in this Light-GBM model, the dataset was separated into training and test datasets. The training dataset, which were used to train the Light-GBM model, contained observations from January 1st, 2019 to December 31st, 2021. The test dataset was used to evaluate the prediction errors of the trained Light-GBM model. 

The second step is the causal discovery phase. To learn the causal structure among the prediction error, prediction target, and covariates in the Light-GBM model, we first evaluated the prediction error in the training dataset with the learned model. After evaluating the prediction error, we use a PCMCI-based algorithm to estimate the causal graph between the prediction error, prediction target, and covariates.

Finally, we conducted the RCA based on the estimated causal graph in the previous step. In this analysis, the root causes of the five largest prediction error sample points were evaluated based on our method and baseline methods to compare the RCA performance for all methods. The five largest prediction error sample points were obtained from the series of prediction errors of the Light-GBM model in test datasets.

The results of this evaluation are shown in Table~\ref{tab:taisetsu_results}, which summarizes the top three highly attributive variables determined by each RCA method (i.e., the selected root-causes).

\begin{table}[h]
    \centering
    \caption{Comparisons of RCA performance in Dam data.}
    \begin{tabular}{c|ccc}
     & 1st variable & 2nd variable & 3rd variable \\
    \hline\hline
    CD-RCA(our) & $q^{(0)}_{0}$ (1.000) & $q^{(2)}_{0}$ (0.369) & $inflow$ (-0.142) \\
    \hline
    LIME & $q^{(1)}_{0}$ (1.000) & $q^{(2)}_{2}$ (-0.799) & $q^{(3)}_{2}$ (-0.600) \\
    \hline
    EIG & $inflow$ (1.000) & $q^{(0)}_{0}$ (0.833) & $q^{(2)}_{0}$ (0.477) \\
    \hline
    LC & $q^{(4)}_{0}$ (1.000) & $q^{(1)}_{0}$ (0.212) & $q^{(3)}_{0}$        (-0.145) \\
    \hline
    GPA & $q^{(1)}_{0}$        (1.000) & $q^{(4)}_{0}$        (0.986) & $q^{(1)}_{1}$        (-0.178) \\
    \hline
    z-score & $q^{(0)}_{0}$ (1.000) & $q^{(2)}_{0}$        (0.942) & $q^{(1)}_{0}$        (0.363) \\
    \end{tabular}
    
    \label{tab:taisetsu_results}
\end{table}
Note that $q_n^{(m)}$ indicates the amount of soil water accumulation in the $n$-th ($n= 0,~1,~2$) tank at the $m$-th observation point ($m = 0,~..,~6$). Moreover, the value in the bracket indicates the estimated value of normalized attribution score in each variable. 

All methods selected one or more of these soil water accumulation variables $q_n^{(m)}$ as top 3 attributive covariates related to prediction error of the dam inflow. Our proposed CD-RCA method identified $q_n^{(0)}$ and $q_n^{(2)}$, which correspond to the $0$-th and $2$-nd observation points in the upper stream of the Ishikari river flowing into the dam (see Fig.~\ref{fig_dam_location}), as the most significant two variables. Among the baselines, only the z-score method could do this. However, the z-score method also identified $q_0^{(1)}$, which does not correspond to the upper stream of the river.  

The result of our CD-RCA method suggests that the anomalous increase in dam inflow may have been caused by abrupt changes in the upstream river (e.g., unexpected strong rain). Since this suggestion is consistent with geographical intuition, our method could help produce interpretable results from RCA analysis in practical situations.

\section{Conclusions} \label{sec:conclusion}
In this study, we tested our proposed CD-RCA method for detecting the root cause variable for outliers in prediction errors of black-box ML systems. By numerical simulations, we confirmed the advantages of our method relative to existing heuristic attribution methods. Moreover, we discovered that the Shapley value might encounter difficulties in detecting the root cause variable when: (1) the amplitude of the root cause $Z$ is lower than that of the observational noise, or (2) there is no causal effect from the root cause variable to the target variable. In future work, we will reveal more rigorously theoretical reasons for these difficulties.  

In addition to the above-mentioned limitation, another issue with the method is that it assumes causal sufficiency in the observational data, meaning no other confounding factors involving the unobserved root cause $Z$. Therefore, future work should consider the effect of unobserved confounding factors for our proposed method. As a possible solution to address this issue, we can use the LPCMCI algorithm \cite{Gerhardus2020_LPCMCI}(PCMCI with latent confounding factors) instead of PCMCI. By using LPCMCI to estimate the causal mechanisms of the prediction error, our proposed method can be used to apply the RCA task under the causal insufficient system. In addition, by combining the LPCMCI with the method for estimating the causal effect boundary \cite{Malinsky2017} in causal insufficient systems, we would sufficiently detect the possible causal structures while considering the distinctiveness issue of the Markov equivalence class under the possibly confounded systems. 

Despite such a limitation, our method provides advantages over the conventional algorithms in the RCA task for understanding the generative mechanisms of the prediction error in ML models. 
\section*{Acknowledgments}
The research was the collaborative work of Toshiba Corporation and Shiga University, based on funding from Toshiba Corporation.

\section*{Code availability statements}
The programming code used to reconstruct all the results of numerical simulations in this paper was implemented
using the language Python and is available at GitHub repository: \url{https://github.com/myGit-YokoyamaHiroshi/CD-RCA_IJCNN2025}

\bibliographystyle{IEEEtran}
\bibliography{0-manuscript.bib}

\appendix 
\renewcommand\thefigure{\thesection.\arabic{figure}}    
\setcounter{figure}{0}  
\setcounter{equation}{0}  
\renewcommand\thetable{\thesection.\arabic{table}}    
\renewcommand{\theequation}{\thesubsection.\arabic{equation}}
\setcounter{table}{0} 

\subsection{Related works}
\label{sec:related_works}
Given its practical importance, there is a vast literature on attribution methods. Some recent attribution methods that do not incorporate causality are~\cite{deng_aaai_2021},~\cite{deng_tpami_2024},~\cite{ide_2021_lc}, and~\cite{ide2023}. Some attribution methods that utilize causality are~\cite{Heskes_neurips_2020},~\cite{do_shapley_2022}, and~\cite{janzing24a}. An application of Shapley-value-based attribution methods is provided in~\cite{makansi2022you}. A related work on feature selection in time-series is~\cite{mastakouri21a}. 

Due to their tractability, additive noise models have been studied extensively. Recent works include causal additive models~\citep{Buhlmann14CAM}, a subclass of additive noise models, and their extensions to cases with unmeasured confounders~\cite{pmlr-v161-maeda21a,pham2025}.

\subsection{Additional materials for experiments}
\subsubsection{Illustrative example}
We use the following linear causal model $F^{train}$:
\begin{align}
\label{eq:illustrative}
    X_1^t &= 0.8 X_1^{t-1} + N_{X_1};\quad X_2^t = 3.8 X_1^{t-1} + 0.8 X_3^{t-1} + N_{X_2};\nonumber \\ 
    X_3^t &= 0.8 X_3^{t-1} + N_{X_3}; \quad
    X_4^t = 3.8 X_2^{t-1} + N_{X_4},
\end{align}
where $N_{X_n} \sim \mathcal{U}(0,~1)$. 

\subsubsection{Effects of outlier magnitude and average total effect}
\label{append1_sim1_equations}
In this section, we explained the details in the equations for generating the synthetic time-series data in the Section~\ref{section_ex_sim1}. As mentioned in this section, we applied two different models $F_1$ and $F_2$ for the exact causal mechanisms in this simulation. The $F_1$ model is:
\begin{align}
\label{eq_sim1_F1}
    X_1^{t} &= 0.8 X_1^{t-1} + N_{X_1}; \ 
    X_2^{t} = 0.8 X_1^{t-1} + 0.8 X_3^{t-1} + N_{X_2};\nonumber \\
    X_3^{t} &= 0.8 X_3^{t-1} + N_{X_3};\ 
    X_4^{t} = 0.8 X_2^{t-1} + N_{X_4}.
\end{align}
The $F_2$ model is:
\begin{align}
\label{eq_sim1_F2}
X_1^{t} &= 0.2 X_1^{t-1} + N_{X_1};\ 
X_2^{t} = 0.2 X_1^{t-1} + 0.4 X_3^{t-1} + N_{X_2};\nonumber \\
X_3^{t} &= 0.2 X_3^{t-1} + N_{X_3}; \ 
X_4^{t} = 0.1 X_2^{t-1} + N_{X_4}.
\end{align}
These models are applied as the causal mechanisms for evaluating the causal attributions of each observed variables for the target anomalous sample.  

The target anomalous sample for the model $F_1$ was generated by the following $F_{1_{test A}}$, $F_{1_{test B}}$, and $F_{1_{test C}}$. The model $F_{1_{test A}}$ is:
\begin{align}
\label{eq_sim1_F1_testA}
    X_1^{t} &= 0.8 X_1^{t-1} + N_{X_1} + Z;\ 
    X_2^{t} = \beta X_1^{t-1} + 0.8 X_3^{t-1} + N_{X_2};\nonumber \\
    X_3^{t} &= 0.8 X_3^{t-1} + N_{X_3};\ 
    X_4^{t} = 0.8 X_2^{t-1} + N_{X_4}.
\end{align}
The model $F_{1_{test B}}$ is:   
\begin{align}
\label{eq_sim1_F1_testB}
&X_1^{t} = 0.8 X_1^{t-1} + N_{X_1};\ 
X_2^{t} = 0.8 X_1^{t-1} + 0.8 X_3^{t-1} + N_{X_2} +\nonumber \\
&Z;\ 
X_3^{t} = 0.8 X_3^{t-1} + N_{X_3};\ 
X_4^{t} = \beta X_2 + N_{X_4}.
\end{align}
The model $F_{1_{test C}}$ is:
\begin{align}
\label{eq_sim1_F1_testC}
&X_1^{t} = 0.8 X_1^{t-1} + N_{X_1};\ 
X_2^{t} = \beta X_1^{t-1} + 0.8 X_3^{t-1} + N_{X_2} +\nonumber\\
&Z;\ X_3^{t} = 0.8 X_3^{t-1} + N_{X_3};
X_4^{t} = 0.8X_2^{t-1} + N_{X_4}.
\end{align}
$N_{X_n} \sim \mathcal{U}(0,~1)$. $\beta$ is selected a number from the interval $[ 0.5,~3.0]$ with $0.2$ step size. $Z$ indicates the root cause factor that follows $Z=z$. The value of $Z=z$ at target sample is selected from the intervals $[0.0,~10]$ with $0.5$ step, otherwise set as $Z=0$. 

The target anomalous sample for the model $F_2$ was generated in the same manner as that of the model $F_1$ by using the following $F_{2_{test A}}$, $F_{2_{test B}}$, and  $F_{2_{test C}}$ models. The model $F_{2_{test A}}$ is: 
\begin{align}
\label{eq_sim1_F2_testA}
    X_1^{t} &= 0.2 X_1^{t-1} + N_{X_1} + Z;\ 
    X_2^{t} = \beta X_1^{t-1} + 0.4 X_3^{t-1} + N_{X_2}\nonumber \\
    X_3^{t} &= 0.2 X_3^{t-1} + N_{X_3};\ 
    X_4^{t} = 0.1 X_2^{t-1} + N_{X_4}.
\end{align}
The model $F_{2_{test B}}$ is:
\begin{align}
\label{eq_sim1_F2_testB}
&X_1^{t} = 0.2 X_1^{t-1} + N_{X_1};\ 
X_2^{t} = 0.2 X_1^{t-1} + 0.4 X_3^{t-1} + N_{X_2} +\nonumber \\
&Z;\ X_3^{t} = 0.2 X_3^{t-1} + N_{X_3};\ 
X_4^{t} = \beta X_2^{t-1} + N_{X_4}.
\end{align}
The model $F_{2_{test C}}$ is:
\begin{align}
    \label{eq_sim1_F2_testC}
&X_1^{t} = 0.2 X_1^{t-1} + N_{X_1};\ 
X_2^{t} = \beta X_1^{t-1} + 0.4 X_3^{t-1} + N_{X_2}+\nonumber \\
&Z;\ X_3^{t} = 0.2 X_3^{t-1} + N_{X_3};
X_4^{t} = 0.1 X_2^{t-1} + N_{X_4}.
\end{align}

\subsubsection{Effects of failed causal graph estimation on CD-RCA}
\label{append_sim2}
The training data in Section~\ref{section_ex_sim2} were generated by the following causal model:
\begin{align}
 \label{eq_sim2_train}
X_1 &= N_{X_1};\ 
X_2 = 3.8 X_1 + 0.8 X_3 + N_{X_2};\nonumber\\
X_3 &= N_{X_3};\ X_4 = 3.8X_2 + N_{X_4},
\end{align}
where $N_{X_n} \sim \mathcal{U}(0,~1)$. $X_4$ is the target of RCA in this simulation.

\end{document}